\documentclass[10pt,twocolumn,letterpaper]{article}
\PassOptionsToPackage{table}{xcolor}
\usepackage[pagenumbers]{cvpr}
\usepackage{multirow}
\usepackage{array}
\usepackage{tabularx}
\usepackage{tikz}
\usetikzlibrary{arrows.meta,positioning,fit,backgrounds,calc}
\definecolor{cvprblue}{rgb}{0.21,0.49,0.74}
\usepackage[pagebackref,breaklinks,colorlinks,allcolors=cvprblue]{hyperref}

\definecolor{pendcolor}{HTML}{C2410C}
\definecolor{ours}{HTML}{2A78D6}

\newcommand{\best}[1]{\textbf{#1}}
\newcommand{\ours}{oracle-supervised critic}
\setlist{itemsep=1pt,topsep=3pt,leftmargin=*}

\def\confName{CVPR}
\def\confYear{2027}

\title{LEDGE: Separating Critic and Editor Failures in Text-to-Image Self-Correction}
\author{
Vyom Agarwal$^{1}$ \quad Mokshda Gangrade$^{1}$ \quad Samet Ayhan$^{1}$ \quad Giacinto Paolo Saggese$^{1}$\\
$^{1}$University of Maryland, College Park \\
{\tt\small \{vyomwal5,mokshdag, sayhan, gsaggese\}@umd.edu}
}

\begin{document}
\maketitle

\begin{abstract}
Text-to-image (T2I) self-correction combines a vision-language model (VLM) critic that diagnoses errors and proposes instructions with an image editor that executes them. End-to-end evaluation obscures whether failures arise from incorrect diagnoses or an editor unable to execute valid instructions. We introduce LEDGE (Learning Error Diagnosis in Generation and Editing), a framework that separates these bottlenecks. LEDGE compiles oracle edit instructions from GenEval while holding the generated images and the editor fixed, isolating the editor's correction capability from the critic's diagnostic accuracy. On Stable Diffusion 3.5 Large, oracle instructions raise GenEval from 72.0\% to 81.2\%, versus 74.1\% for the baseline VLM critic. Yet 435 of 644 failed images remain unrepaired, exposing a substantial editor gap. We also use oracle labels from disjoint training prompts to supervise the critic. Oracle supervision raises exact diagnosis from 33.3\% (the same VLM with our structured output, untrained) to 70.8\%. With the same editor, the resulting critic improves GenEval from 42.2\% (uncorrected) to 59.4\% on Stable Diffusion 1.5, against 50.9\% for the baseline critic, and from 72.0\% to 76.0\% on Stable Diffusion 3.5 Large, despite training only on Stable Diffusion 1.5. Improvements transfer to unseen object classes, a second editor, and independent T2I-CompBench judges and prompts. 
A small human evaluation confirms the ranking in estimated net corrections but suggests a smaller gain than detector-based evaluation.
LEDGE thus distinguishes two limits of self-correction: knowing what to fix and being able to fix it.\end{abstract}
\section{Introduction}

Text-to-image (T2I) models still fail on compositional prompts: they drop objects, miscount, color the wrong object and ignore spatial relations. On GenEval~\citep{ghosh2023geneval}, Stable Diffusion~1.5~\citep{rombach2022ldm} places objects correctly in 4.5\% of position prompts and binds colors correctly in 7.8\% of color-attribution prompts. A common remedy is \emph{self-correction}: a vision-language model (VLM) critic inspects the generated image and writes an editing instruction, which an image editor then executes~\citep{r3bench2026,selfcorrected2026,rpo2025,jaiswal2026iterative,chu2026visiondirector}.

When such a loop fails, there are two possible causes. The critic may misdiagnose the image or phrase an instruction that the editor cannot use. Alternatively, the editor may fail to carry out a valid correction. Prior work has established that VLM guidance improves editing and that critic-driven refinement improves generation~\citep{fu2024mgie,jaiswal2026iterative}, but it evaluates the correction process as a whole, so these two failures cannot be told apart. This distinction is crucial: improving the editor cannot compensate for incorrect instructions, just as improving the critic cannot help if the editor fails to execute them.

In this paper, we introduce LEDGE (Learning Error Diagnosis in Generation and Editing), which separates the two causes using a simple observation. Benchmarks such as GenEval specify each prompt's requirements in structured form and report which requirement a failed image violates. From these we compile \emph{oracle instructions}, i.e., the correct edit for every failure, phrased with the same templates used to render the critic's instructions (Figure~\ref{fig:pipeline}). With the editor and the images held fixed, the oracle measures how much the instruction alone limits correction (the \emph{instruction gap}) and how much remains even with correct instructions (the
\emph{editor gap}).

The oracle cannot be used at test time, since it reads the evaluator's verdict, but the diagnosis it enables suggests a direct use: it provides free supervision. To this end, we generate images on training prompts whose specifications are disjoint from the benchmark's, label them with the oracle, and fine-tune the baseline model. The resulting oracle-supervised critic then lists every requirement violation in a structured form, from which editing instructions are rendered deterministically. On SD1.5 with the VAREdit editor~\citep{varedit2025}, the oracle-supervised critic closes most of the instruction gap with fewer regressions than the baseline critic, and, although trained only on SD1.5 images, it also improves correction of SD3.5-Large images, so the learned diagnosis transfers across generators. The ranking of critics also holds when the same images are re-scored by the independent judges of T2I-CompBench~\citep{huang2023t2icompbench}.

Our contributions are summarized as follows:
\begin{itemize}
\item We introduce LEDGE, which holds the images and the editor fixed and compiles oracle instructions from GenEval's verdicts. With either of two editors, oracle instructions repair $2.3\times$ more SD1.5 failures than the baseline critic (Qwen3-VL-8B)~\citep{qwen3vl2025}, yet 61\% of SD1.5 and 68\% of SD3.5 failures remain unrepaired (Section~\ref{sec:gap}).
\item We analyze why the baseline critic fails: it identifies the right error in 39\% of failures, defers 35\% to review and misses 12\%, and on counting it often asks to remove objects that are missing (Section~\ref{sec:critic-failures}).
\item We propose oracle-supervised critics, trained by LoRA on labels compiled automatically from prompts disjoint from the benchmark. They raise exact diagnosis from 33.3\% to 70.8\% and GenEval from 50.9\% to 59.4\% with the same editor, and the same supervision improves a Molmo critic from 52.4\% to 56.5\% (Section~\ref{sec:critic}).
\item We show that the critic, trained only on SD1.5 images, transfers to SD3.5-Large (72.0\% to 76.0\%), to object classes absent from training, and to a second editor, FLUX.1~Kontext. The ranking of critics also holds under T2I-CompBench's independent judges, where the oracle-supervised critic recovers 97\% of the oracle's gain (Section~\ref{sec:generalization}).
\end{itemize}

\section{Related Work}

\paragraph{VLM-guided correction.}
Multimodal LLMs already guide instruction-based editing: MGIE~\citep{fu2024mgie} learns expressive instructions that steer a diffusion editor end to end, and SmartEdit~\citep{huang2024smartedit} uses an MLLM to interpret complex instructions. For T2I generation, a VLM critic drives iterative refinement of the generator's output~\citep{jaiswal2026iterative}, rewrites the prompt between rounds~\citep{khan2025tir,dong2025lumigen}, decomposes long instructions into verified, reversible edits~\citep{chu2026visiondirector}, or localizes distortions for repair~\citep{shen2026retoucher}. EditRefiner~\citep{xu2026editrefiner} refines edited images with an agentic loop, and Inline Critic~\citep{kang2026inlinecritic} predicts errors inside the editor's forward pass. R3-Bench~\citep{r3bench2026} scores a VLM's verdict, critique and rectification of T2I outputs, and self-corrected image generation~\citep{selfcorrected2026} closes the loop at generation time. These works establish that critic-guided correction helps, and we do not claim the loop itself as a contribution. They evaluate it as a whole; LEDGE instead holds the images and the editor fixed to separate diagnosis from execution, and uses the resulting oracle to supervise the critic.

\paragraph{Improving editing instructions.}
EditThinker~\citep{editthinker2025} trains a Qwen3-VL-8B model that critiques an editor's output, rewrites the instruction and retries, targeting user editing requests on editing benchmarks. Complex-Edit~\citep{complexedit2025} studies how instruction complexity affects editors. Our setting differs in two ways: the target is fidelity to the original T2I prompt, and our supervision is compiled from a verifier rather than generated as reasoning data.

\paragraph{Instruction-guided editors.}
Editors are trained on synthetic or curated pairs~\citep{brooks2023instructpix2pix,zhang2023magicbrush,hui2025hqedit},
with human feedback~\citep{zhang2024hive} and, more recently, with online reinforcement learning
against learned reward models~\citep{editscore2026,uniworldv2,verifierrl2026,spatialreward2026};
training-free methods regulate the editing strength during denoising~\citep{fu2025feededit}.
Editing benchmarks evaluate editors on curated instructions~\citep{basu2023editval,jia2026compbench}, while MLLMs can detect and assess the changes an edit makes~\citep{baraldi2025dice}. In contrast, we do not train the editor. We hold it fixed (VAREdit~\citep{varedit2025} and, as a second editor, FLUX.1~Kontext~\citep{kontext2025}) and evaluate it on the instructions required to correct a generator's actual failures, thereby measuring the editor gap.

\paragraph{Training generators from corrections and rewards.}
RPO~\citep{rpo2025} turns VLM critiques into edits and trains SD1.5 and SDXL with
Diffusion-DPO~\citep{wallace2024diffusiondpo} on (original, edited) pairs, and region-aware
bimodal DPO~\citep{regiondpo2026} builds edit-based preference pairs for compositional generation.
Online reinforcement learning with verifiable GenEval-style rewards~\citep{flowgrpo2025,dgpo2026}
reaches high GenEval scores on prompts generated from the benchmark's own templates, and
Diffusion-SDPO~\citep{sdpo2025} addresses objectives that improve by degrading the losing sample.
Unlike these methods, we train the critic rather than the generator and keep the editor fixed, so
that gains can be attributed to the instruction alone.

\paragraph{Evaluation of compositional generation.}
GenEval~\citep{ghosh2023geneval} checks object presence, counts, colors and relations with a
Mask2Former detector~\citep{cheng2022mask2former} and a CLIP~\citep{radford2021clip} color
classifier. TIFA~\citep{hu2023tifa} and VQAScore~\citep{lin2024vqascore} score faithfulness with visual question answering. Other benchmarks use different judges, including T2I-CompBench~\citep{huang2023t2icompbench} (unrelated to the editing benchmark CompBench),
GenEval~2~\citep{geneval2}, RichHF-18K~\citep{liang2024richhf} and EvalMuse-40K~\citep{evalmuse2024}.
We build the oracle on GenEval and re-score its images with T2I-CompBench's judges as a check on its detector (Section~\ref{sec:generalization}).

\section{Problem Setup}

\begin{figure*}[t]
\centering
\resizebox{0.86\linewidth}{!}{%
\begin{tikzpicture}[
  box/.style={draw=black!60, rounded corners=2pt, fill=white, align=center, font=\small,
              minimum height=1.0cm, minimum width=2.2cm, inner sep=3pt},
  model/.style={box, fill=black!5},
  key/.style={box, fill=ours!10, draw=ours},
  orc/.style={box, fill=pendcolor!8, draw=pendcolor!70, minimum width=3.4cm},
  arr/.style={-{Stealth[length=2.2mm]}, thick, black!70},
  darr/.style={-{Stealth[length=2.2mm]}, thick, dashed, black!55},
  lab/.style={font=\footnotesize, black!75, align=center}]
\node[box]   (p)  at (0,0)     {Prompt $p$};
\node[model] (g)  at (3.0,0)   {Generator $G$\\[-1pt]\scriptsize SD1.5 / SD3.5};
\node[box]   (x)  at (6.0,0)   {Image $x$};
\node[key]   (c)  at (9.3,0)   {Critic $C$\\[-1pt]\scriptsize VLM: list violations};
\node[box]   (r)  at (12.8,0)  {Renderer\\[-1pt]\scriptsize fixed templates};
\node[orc]   (o)  at (9.3,2.3) {Oracle compiler\\[-1pt]\scriptsize requirements + detector reasons};
\node[model] (e)  at (12.8,-2.3) {Editor $E$\\[-1pt]\scriptsize VAREdit / Kontext};
\node[box]   (xp) at (9.3,-2.3)  {Edited image $x'$};
\node[model] (v)  at (6.0,-2.3)  {Verifier $V$\\[-1pt]\scriptsize GenEval};
\draw[arr] (p) -- (g);
\draw[arr] (g) -- (x);
\draw[arr] (x) -- (c);
\draw[arr] (c) -- node[lab, below]{edits} (r);
\draw[arr] (r) -- node[lab, right]{instruction $I$} (e);
\draw[arr] (e) -- (xp);
\draw[arr] (xp) -- (v);
\draw[darr] (x.north) |- (o.west);
\draw[darr] (o.east) -| node[lab, pos=0.75, right]{oracle\\edits $I^\star$} (r.north);
\draw[darr, ours] (o.south) -- node[lab, right, text=ours]{oracle labels on\\\emph{training} prompts\\$\rightarrow$ LoRA} (c.north);
\end{tikzpicture}}
\caption{Overview of LEDGE. The critic lists requirement violations; a fixed renderer turns them into an
instruction for a fixed editor. The oracle compiler produces the correct edits from the benchmark's
requirements and the detector's failure reasons. On test images it is a reference; on training
prompts it supervises the critic (blue).}
\label{fig:pipeline}
\end{figure*}
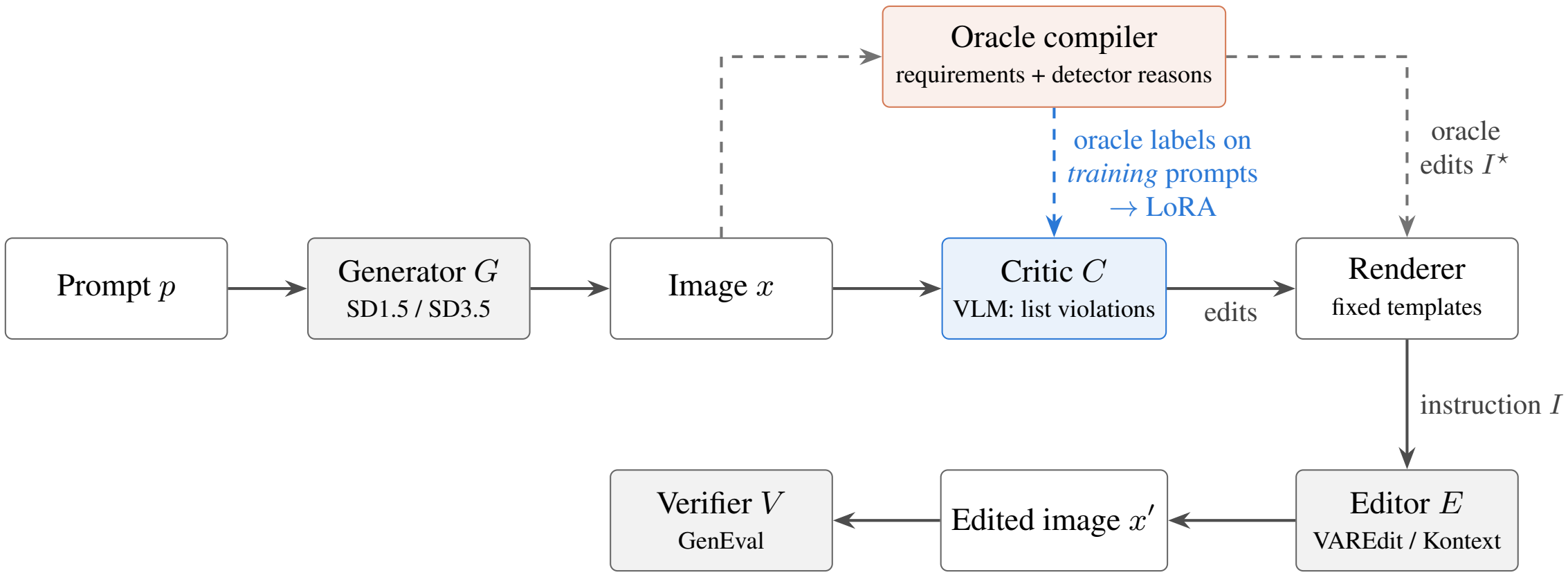

\begin{sloppypar}
We first define the correction loop and the quantities that LEDGE measures. A generator $G$ produces an
image $x$ for prompt $p$. A critic $C$ returns a decision
$d\in\{\textsc{keep},\textsc{edit},\textsc{review}\}$ and, for \textsc{edit}, an instruction $I$.
An editor $E$ produces $x'=E(x,I)$, while \textsc{keep} and \textsc{review} retain $x$. A verifier
$V$ returns $V(x,p)\in\{0,1\}$ (Figure~\ref{fig:pipeline}).
\end{sloppypar}

\paragraph{Repairs and regressions.} For each image, a \emph{repair} is $V(x,p)=0 \wedge V(x',p)=1$
and a \emph{regression} is $V(x,p)=1 \wedge V(x',p)=0$. We report both and test paired differences
with an exact McNemar test~\citep{mcnemar1947,dietterich1998} on the discordant images, i.e., a
two-sided binomial test of repairs against regressions.
\paragraph{Scores.} GenEval's overall score is the \emph{macro} average of its six task accuracies
(single object, two objects, counting, colors, position and color attribution); the \emph{micro}
score is the fraction of all images that pass. Repairs and regressions convert directly to micro,
whereas macro weights each task equally (task sizes range from 320 to 400 images; see
Appendix~\ref{app:macro}).

\paragraph{Instruction and editor gaps.} Let $S(I)$ denote the score after editing with
instructions $I$, $I_C$ the critic's instructions and $I^\star$ the oracle's. The
\emph{instruction gap} is $S(I^\star)-S(I_C)$, and the \emph{editor gap} is the fraction of failures
that remain unrepaired under $I^\star$. For an improved critic $C'$, the share of the gap closed is
$\boldsymbol{\big(S(I_{C'})-S(I_C)\big)/\big(S(I^\star)-S(I_C)\big)}$.

\section{Oracle Instructions}
\label{sec:oracle}

GenEval stores each prompt's requirements (objects, counts, colors and relations to other objects)
and, for every image, the detector's reasons for failure. We parse five reason types: an object is
\emph{missing} (fewer instances than required), \emph{excess} (more than allowed, which applies to
counting prompts only), \emph{wrong color}, \emph{wrong relation}, or \emph{no target} (the
reference object of a relation is absent). Every failure on the four generators we tested (3,802
images) parses without exception.

\paragraph{From diagnosis to instruction.} A correct diagnosis is not yet an executable
instruction. GenEval's reason ``expected handbag$\geq$4, found 2'' says what is wrong, but an editor
also needs to know how many objects to add, where to place them and which total to keep. The
compiler therefore maps each reason to a structured edit whose arguments come from the prompt's
requirements. A \emph{missing} object becomes an \texttt{add} with the number to add and the total to
leave (``Add two more handbags \ldots, leaving exactly four handbags''), placed on the correct side
of its relation partner (``Add one computer keyboard above the zebra'') and added together with the
partner when both are missing. \emph{Excess} becomes a \texttt{remove} with the total to keep,
\emph{wrong color} a \texttt{recolor} of every instance, \emph{wrong relation} a \texttt{move} of the
object that carries the relation, and \emph{no target} is resolved by adding the missing reference.
Excess counts are first checked for duplicate detections: if overlapping boxes (IoU $>0.5$) explain
the excess, the image is sent to \textsc{review} rather than told to delete a real object (3 of the
3,802 failures). Edits are emitted in a fixed order (additions, removals, recolors, moves) and
rendered with the same templates as the critic (Appendix~\ref{app:templates}), so the oracle and the
structured critic differ only in their diagnoses. The oracle's instruction is one valid edit among
several (moving the reference object would also fix a relation), so it is a reference for this
instruction format rather than a bound over all possible instructions.

Note that the oracle reads the evaluator's verdict. It is thus a reference for analysis and
never a test-time method. It also inherits the evaluator's errors, which we examine with human
inspection and with independent judges in Section~\ref{sec:generalization}.

\section{Oracle-Supervised Critics}
\label{sec:method}

This section describes the critic's output format, its training data, and how we train and select it. We refer to the original Qwen3-VL-8B-Instruct model as the \emph{baseline critic}, its zero-shot variant with structured output as the \emph{structured zero-shot critic}, and the LoRA-fine-tuned variant as the \emph{oracle-supervised critic}.

\paragraph{Structured critique.} The baseline critic is prompted to return one
actionable error from four types, or \textsc{keep}/\textsc{review}. However, more than a quarter of our training failures (367 of 1,339) require two edits, including about half of the counting failures (Table~\ref{tab:labels}). A single-error format cannot capture these cases. Both structured critics instead return
\texttt{\{"decision": keep|edit, "edits": [...]\}}, where each edit is an \texttt{add},
\texttt{remove}, \texttt{recolor} or \texttt{move} record with its object, count, color, and relation fields (Appendix~\ref{app:schema}). The editing instructions are rendered from the structured edits using fixed templates. Applying the same rendering process to oracle edits reproduces their instructions exactly for all 3,799 failures that receive edit lists (the remaining 3 of the 3,802 are sent to \textsc{review}).

\paragraph{Training data.} We use a pilot set of GenEval-style prompts covering 20 COCO classes. It contains 765 training
prompts and 188 development (DEV) prompts, with two SD1.5 samples each (1,530 and 376 images). The
20 classes appear in GenEval's vocabulary, but 65\% of the official test images (1,428 of 2,212)
involve none of them.
Before labeling, we correct the pilot metadata to the official conventions. The original export
placed an upper-count constraint on every prompt and stored relations with forward indices; we
restrict upper counts to counting prompts (affecting 577 of 765 training prompts) and re-score the
existing images without regeneration. The oracle compiler then labels these training images: 1,339
failures receive edit lists and 191 correct images receive \textsc{keep}.

\paragraph{Train/test separation.} Oracle labels supervise the critic only on the pilot prompts.
No training or DEV prompt shares a test prompt's specification (task, objects, counts, colors and
relation), the pilot images are generated separately from the test images, and checkpoints are
selected on DEV, so no test prompt, image or verdict is used for training or selection. Because
GenEval prompts follow fixed templates, disjoint specifications do not rule out shared templates; we
therefore also report results on the test images whose objects never appear in training
(Section~\ref{sec:generalization}).

\paragraph{Fine-tuning.} We train LoRA adapters~\citep{hu2022lora} on the language model of
Qwen3-VL-8B, with the vision encoder frozen. The model is trained to emit the target JSON given the image, the original prompt, and a rules prompt. Correct images are oversampled $3\times$ to counter the 12.5\% keep rate.
Hyperparameters are given in Table~\ref{tab:hparams}; checkpoints are saved every 50 optimizer
steps (717 in total).

\paragraph{Checkpoint selection.} The selection rule is fixed in advance. On DEV, we choose the checkpoint with the highest exact-diagnosis rate, subject to editing at most 10\% of correct DEV images. Ties are resolved in favor of the earlier checkpoint.  The official benchmark is used once, with the
selected checkpoint (Appendix~\ref{app:checkpoints}).

\section{Experiments}

We use LEDGE to evaluate three claims. First, both an instruction gap and an editor gap exist and can be measured separately (Section~\ref{sec:gap}). Second, oracle supervision closes much of the instruction gap (Sections~\ref{sec:critic-failures} and~\ref{sec:critic}). Third, the oracle-supervised critic generalizes beyond its training conditions (Section~\ref{sec:generalization}).
\begin{table}[t]
\centering
\caption{GenEval on SD1.5 images with the same VAREdit editor and the same 2,212 images. Rep./Reg.:
repairs and regressions against the original images; $p$: exact McNemar test against the baseline
critic on the same images. Best non-oracle results in bold.}

\label{tab:main}
\small
\setlength{\tabcolsep}{3.5pt}
\begin{tabular}{lrrrrl}
\toprule
Instructions from & Macro & Micro & Rep. & Reg. & $p$ \\
\midrule
None (original) & 42.2 & 40.1 & -- & -- & -- \\
Generic instruction & 49.9 & 48.4 & 246 & 63 & $0.33$ \\
Baseline (Qwen) & 50.9 & 49.3 & 226 & 23 & -- \\
Molmo + rules & 52.4 & 50.9 & 301 & 64 & $0.09$ \\  
Molmo, oracle-sup. & 56.5 & 55.2 & 345 & 12 & $7{\times}10^{-13}$ \\
\textcolor{ours}{\textbf{Ours}} (oracle-sup.) & \best{59.4} & \best{58.3} & \best{420} & \best{18} & $6{\times}10^{-28}$ \\
\midrule
Oracle (reference) & 64.3 & 63.3 & 512 & 0 & $1{\times}10^{-56}$ \\
\bottomrule
\end{tabular}
\end{table}
\subsection{Experimental Setup}

\paragraph{Generators and images.} SD1.5 at $512^2$ and SD3.5-Large at $1024^2$ generate four
samples for each of GenEval's 553 prompts (2,212 images). SDXL~\citep{podell2023sdxl} and
FLUX.1-dev~\citep{flux2024} serve as reference points (Table~\ref{tab:generators}). We verify that
the SD1.5 images used in every comparison match their generation-time SHA-256 checksums (2,212 of
2,212) and that the run manifest matches the hash recorded when the baseline was scored.

\paragraph{Editors.} VAREdit-8B at 512px~\citep{varedit2025} (CFG 3.0, $\tau=0.1$, seed 42) is the
main editor. FLUX.1~Kontext [dev]~\citep{kontext2025} (28 steps, guidance 2.5, seed 42) is the
second editor (Section~\ref{sec:generalization}). Every edited output is stored at the source resolution.

\paragraph{Resolution control.} VAREdit-8B outputs $512^2$ images. For 1024px generators we
thus create a Lanczos-downscaled $512^2$ copy of the run. We use this copy for the baseline, all critics, the oracle, and the editor, ensuring that kept and edited images share the same resolution. The downscaled SD3.5 copy scores 72.0\%, compared with 71.0\% at native resolution.

\paragraph{Critics.} We compare (i) the baseline critic (\emph{Qwen3-VL-8B} with single-error prompting);
(ii) the \emph{structured zero-shot critic}, the same model with our structured format and no training;
(iii) the \emph{\ours}, i.e., (ii) after LoRA training; (iv) \emph{Molmo + rules}, in which
Molmo~\citep{molmo2} is prompted zero-shot to report the observed objects, counts, colors and
relations and a rule script converts them into an instruction; (v) \emph{generic}, which edits every
image with a fixed instruction that quotes the original prompt, asks the editor to make the image
satisfy it, and asks it to leave a correct image unchanged; and (vi) the oracle.

\paragraph{Statistics.} Paired differences use exact McNemar tests. To estimate variability, we bootstrap over prompts, stratified by task. The SD1.5 baseline has a standard error of 1.00 points macro (0.98 points micro) and a 95\% confidence interval of 40.3--44.2\%.

\subsection{The Instruction Gap}
\label{sec:gap}

\begin{figure}[t]
\centering
\includegraphics[width=\linewidth]{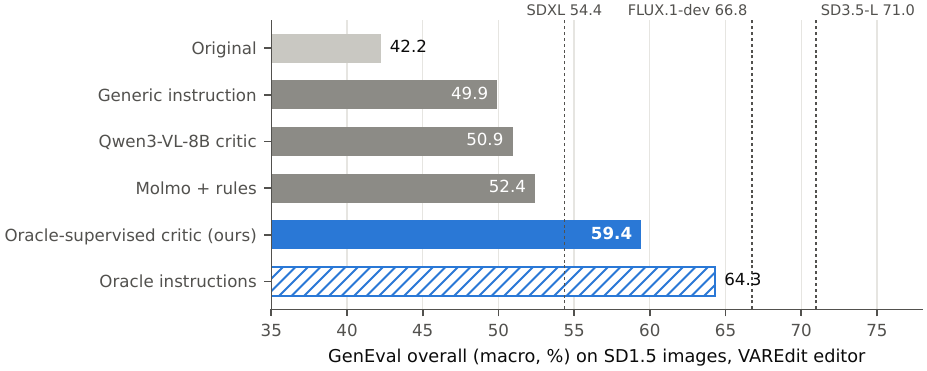}
\caption{GenEval overall score on SD1.5 images by instruction source, same editor. Dashed lines mark
larger generators without correction. With the oracle-supervised critic, SD1.5 plus correction
exceeds SDXL; whether this holds at matched compute is not established.}
\label{fig:main}
\end{figure}

As shown in Table~\ref{tab:main} and Figure~\ref{fig:main}, oracle instructions repair 512 of 1,324
failures with the editor held fixed and raise GenEval from 42.2\% to 64.3\%. The baseline critic repairs 226 failures and reaches 50.9\%, capturing 39\% of the oracle's gain over the original
images. The instruction gap is therefore 13.4 points.

A generic instruction repairs \emph{more} failures than the baseline critic (246 vs.\ 226) but breaks nearly
three times as many correct images (63 vs.\ 23). This suggests that the baseline critic's advantage over the
generic prompt comes from abstaining on correct images rather than from more useful instructions;
paired on the same images, the two do not differ significantly (177 vs.\ 197 images; $p=0.33$).

\begin{figure}[t]
\centering
\includegraphics[width=\linewidth]{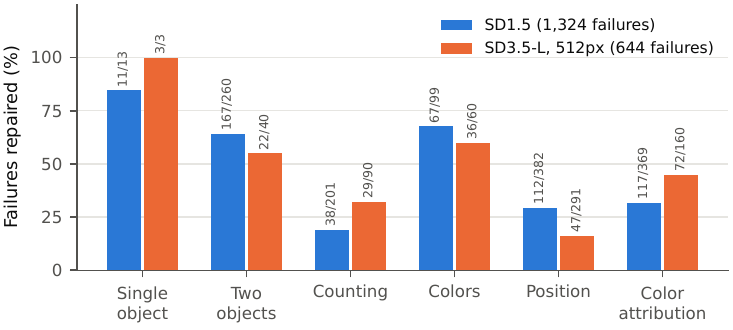}
\caption{Editor gap of VAREdit. Bars give the share of failures repaired when VAREdit receives
\emph{oracle} instructions; labels give repairs / failures per task.}
\label{fig:editor-gap}
\end{figure}

\paragraph{The editor gap.} Even with oracle instructions, 812 of 1,324 SD1.5 failures (61.3\%)
and 435 of 644 SD3.5 failures (67.5\%) remain unrepaired (Figure~\ref{fig:editor-gap}). Counting is
the hardest task, with 38 of 201 failures (19\%) repaired, followed by position (29\%) and color
attribution (32\%); on SD3.5, the repair rate for position falls to 16\%. To see how the editor
fails, we compare GenEval's failure reasons before and after each oracle edit. Among the unrepaired
SD1.5 images, an object the instruction asks to add is still missing in 56.3\% (failed
insertion), a recolored object keeps or receives the wrong color in 17.5\% (incorrect attribute
modification), a moved or added object is still misplaced in 18\%, and a requirement that held
before editing is now violated in 9.5\% (unintended change); an image can fall into several
categories (Appendix~\ref{app:editor-failures}). On a 600-image sample, regenerating each failure
once with a new seed repairs 16\% of failures against 37\% for oracle-instructed editing, but as many
on counting (15 vs.\ 13 of 58), where the editor is thus the bottleneck (Appendix~\ref{app:resample}). Out of the 420 repairs of the \ours, 30 (7\%) are not repaired by the oracle, against 56 of 226 (25\%) for the baseline critic and 81 of 246 (33\%) for the generic instruction.

\subsection{Why the Baseline Critic Fails}
\label{sec:critic-failures}

\begin{table*}[t]
\centering
\caption{Baseline critic decisions against GenEval's verdict on the original SD1.5 images (left) and its
diagnoses of failures against the oracle (right).}
\label{tab:qwen-failures}
\small
\begin{tabular}{lrr}
\toprule
Qwen decision & Correct (888) & Failed (1,324) \\
\midrule
\textsc{keep} & 789 (88.9\%) & 164 (12.4\%) \\
\textsc{edit} & 65 (7.3\%) & 698 (52.7\%) \\
\textsc{review} & 34 (3.8\%) & 462 (34.9\%) \\
\bottomrule
\end{tabular}
\hspace{1.2em}
\begin{tabular}{lr}
\toprule
Outcome on 1,324 failures & Share \\
\midrule
Right error type and object & 39.4\% \\
Sent to review & 34.9\% \\
Kept (error missed) & 12.4\% \\
Wrong error type & 10.4\% \\
Right type, wrong object & 2.9\% \\
\bottomrule
\end{tabular}
\end{table*}

Table~\ref{tab:qwen-failures} shows that the baseline critic is a good gate but a weak diagnoser. It keeps 89\% of correct images but edits only 53\% of failures, correctly identifying the error and object in 39\% of failed images. Position is the hardest category to diagnose, with 58\% of position failures
sent to review. All 23 regressions come from the 65 correct images that the baseline critic chose to edit, a
regression rate of 35\% on those images. On counting failures where the detector found \emph{too
few} objects, the baseline critic asks to remove extras 84 times and to add objects 33 times. Part of this
may reflect detector undercounts at GenEval's 0.9 counting threshold.

\subsection{Oracle-Supervised Critic Results}
\label{sec:critic}

\begin{table}[t]
\centering
\caption{Diagnosis on official SD1.5 images (888 correct, 1,324 failed) against the oracle, without
editing. \emph{Exact}: same edit set as the oracle; \emph{Any}: at least one matching edit
($^\dagger$type-level only, not directly comparable). The gray row ignores the image and is tuned on
the test failures, an upper bound for prompt-only critics. Best critic results in bold.}
\label{tab:diagnosis}
\small
\setlength{\tabcolsep}{3pt}
\begin{tabular}{lrrrrr}
\toprule
& \multicolumn{2}{c}{Correct images} & \multicolumn{3}{c}{Failed images} \\
\cmidrule(lr){2-3}\cmidrule(lr){4-6}
Critic & Keep & Edit $\downarrow$ & Edit & Exact & Any \\
\midrule
Baseline (single error) & 88.9 & 7.3 & 52.7 & -- & 42.3$^\dagger$ \\
Structured zero-shot & 77.5 & 22.0 & 91.4 & 33.3 & 43.6 \\
\textcolor{ours}{\textbf{Ours}} (oracle-sup.) & \best{93.5} & \best{6.5} & 89.0 & \best{70.8} & \best{79.3} \\
\midrule
\rowcolor{gray!20} Prompt-only rule & 0.0 & 100.0 & 100.0 & 49.1 & 57.8 \\
\bottomrule
\end{tabular}
\end{table}

\paragraph{Diagnosis.} Fine-tuning raises exact diagnosis from 33.3\% to 70.8\% (Table~\ref{tab:diagnosis}) and lowers edits on correct images from 22.0\% to 6.5\%, below the baseline critic's 7.3\%. The structured format alone brings no clear gain: zero-shot, it edits 22.0\% of correct images, three times the baseline critic's rate, and its any-match accuracy (43.6\%) is close to the baseline's 42.3\%, a looser type-level match. The gain thus comes from training.

\paragraph{Does the critic use the image?} GenEval prompts follow fixed templates, so a critic might predict errors from the prompt alone. A prompt-only rule that outputs each task's most common edit pattern, chosen on the test failures, reaches 49.1\% exact diagnosis (Table~\ref{tab:diagnosis}) but must edit every correct image; over all 2,212 images it is right on 29.4\%, against 79.9\% for the \ours\ and 51.0\% for the structured zero-shot critic. Our critic exceeds the rule on position (84\% vs.\ 44\%), color attribution (58\% vs.\ 29\%) and two objects (78\% vs.\ 59\%), where the image decides the error, and trails it on counting (73\% vs.\ 76\%) and colors (51\% vs.\ 60\%), where one error type dominates.

By task (Figure~\ref{fig:diagnosis} in the supplement), exact diagnosis rises from 40\% to 84\% on position, 13\% to 58\% on color attribution and 36\% to 78\% on two objects, although training contains only 19 two-object prompts. Colors barely improves (49\% to 51\%); the critic keeps 39 of its 99 failures.

\paragraph{End-to-end correction.} With the same editor, the \ours\ raises GenEval to 59.4\% macro
and 58.3\% micro (Table~\ref{tab:main}), with 420 repairs and 18 regressions. It recovers 63\% of
the instruction gap (8.5 of 13.4 points) and 82\% of the oracle's repairs. Compared with the baseline, 273 images pass only under our critic, versus 74 passing only under the baseline ($p=6\times10^{-28}$), for a gain of 8.5 points macro.

\begin{figure*}[t]
\centering
\begin{subfigure}[t]{0.49\linewidth}\centering
\includegraphics[width=\linewidth]{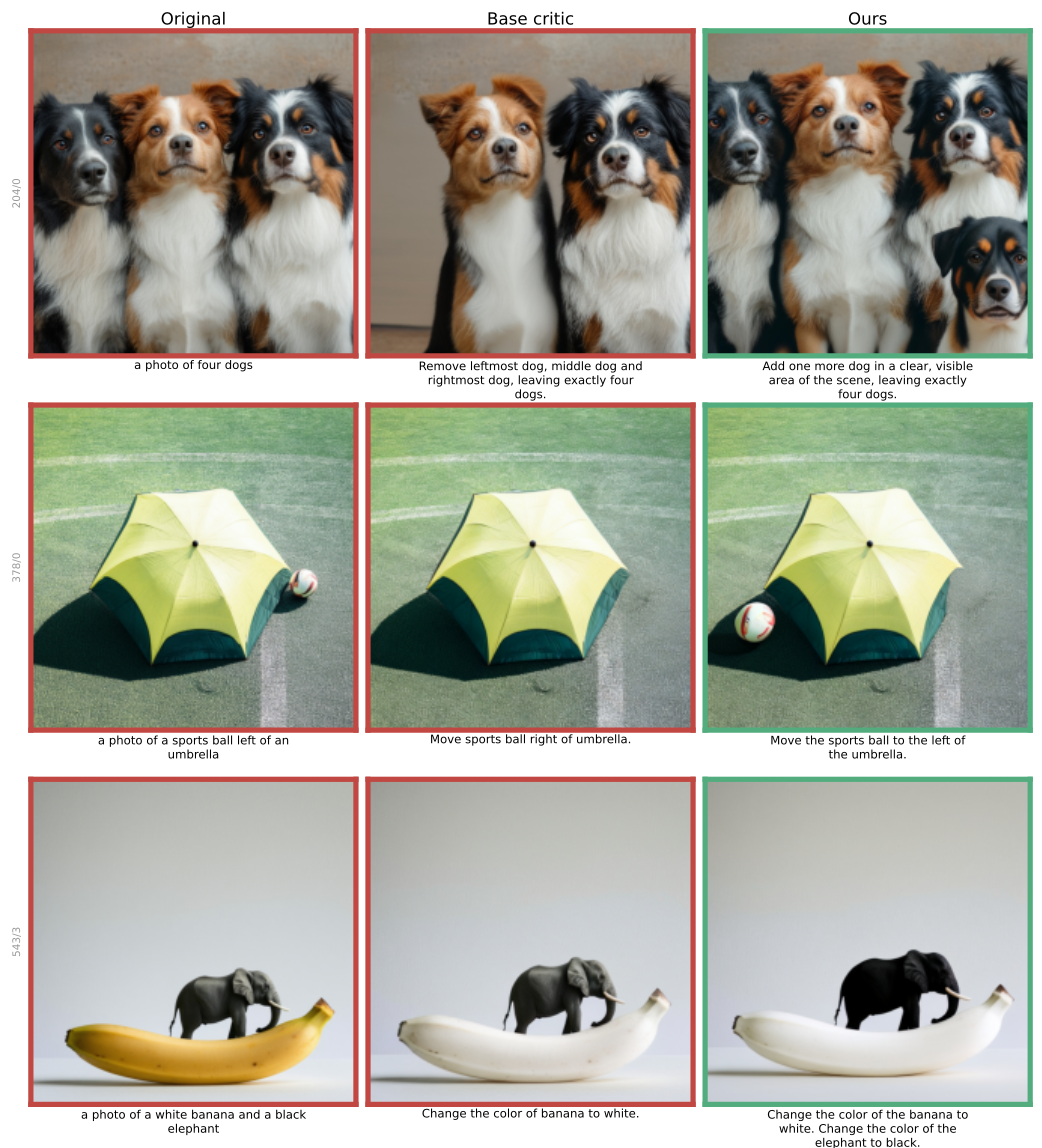}\caption{SD3.5-Large}\end{subfigure}\hfill
\begin{subfigure}[t]{0.49\linewidth}\centering
\includegraphics[width=\linewidth]{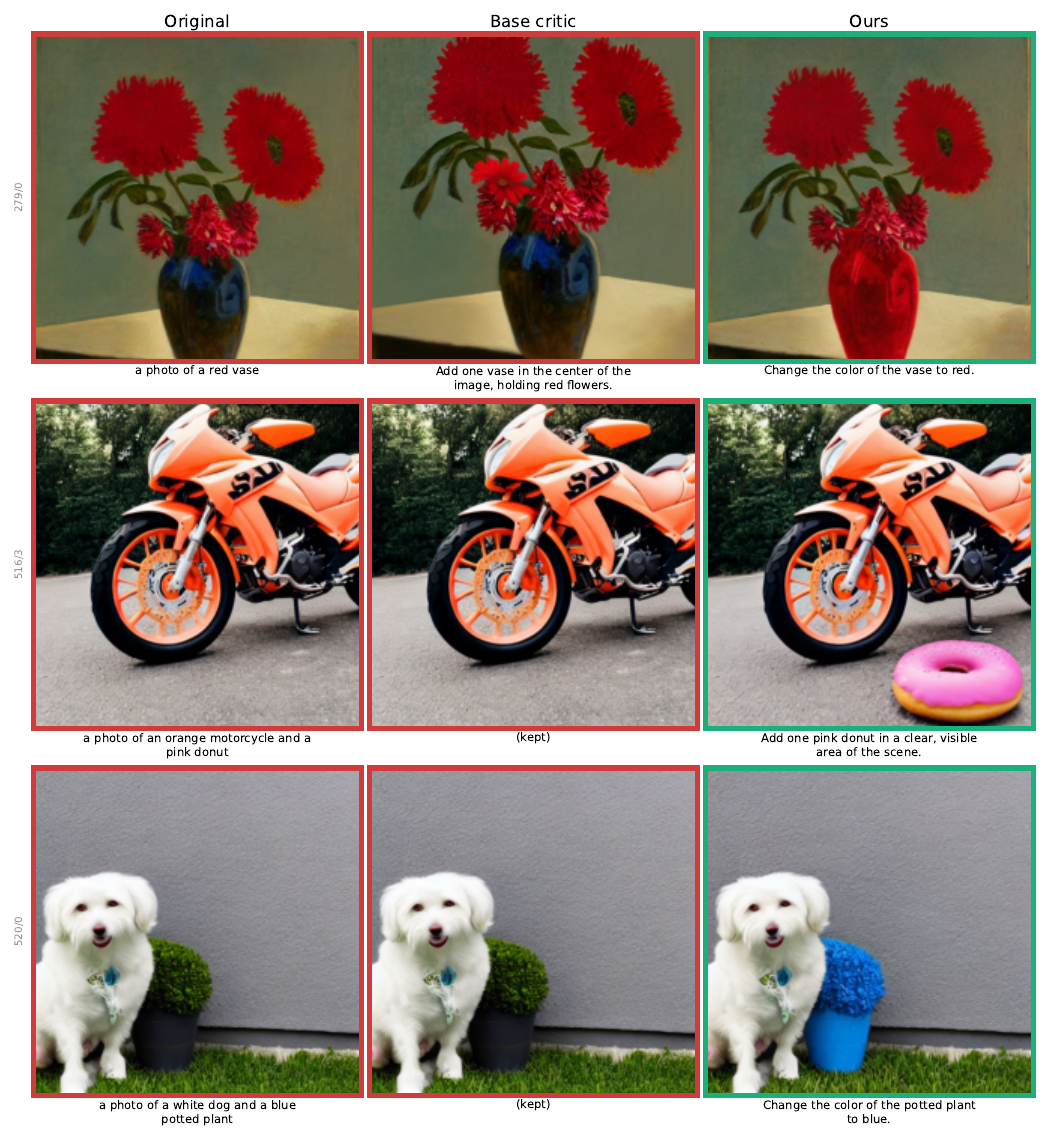}\caption{SD1.5}\end{subfigure}
\caption{Examples where the baseline critic's correction fails and the oracle-supervised critic
succeeds (same editor). Each row is one example; columns show the original image, the baseline
critic's result and ours, with GenEval's verdict as the frame color (green: pass, red: fail) and
the instruction below each edited image.
Examples are hand-selected from the images whose original fails GenEval, whose baseline-critic correction still fails and whose oracle-supervised correction passes (258 on SD1.5, 63 on SD3.5).}
\label{fig:contrast}
\end{figure*}

\begin{table}[t]
\centering
\caption{Per-task accuracy on SD1.5 after editing. Qwen: baseline critic; Molmo: Molmo + rules;
closed: share of the baseline-to-oracle gap closed by the oracle-supervised critic. Bold marks the
best non-oracle result for counting and for the macro average.}
\label{tab:per-task}
\small
\setlength{\tabcolsep}{3.5pt}
\begin{tabular}{lrrrrrr}
\toprule
Task & Orig. & Qwen & Molmo & Ours & Oracle & Closed \\
\midrule
Single object & 95.9 & 96.6 & 95.0 & 97.8 & 99.4 & 44\% \\
Two objects & 34.3 & 58.1 & 53.0 & 69.7 & 76.5 & 63\% \\
Counting & 37.2 & 40.0 & \best{45.0} & 43.8 & 49.1 & 41\% \\
Colors & 73.7 & 81.9 & 78.7 & 84.0 & 91.5 & 22\% \\
Position & 4.5 & 10.8 & 28.5 & 29.8 & 32.5 & 87\% \\
Color attr. & 7.8 & 18.3 & 14.3 & 31.5 & 37.0 & 71\% \\
\midrule
Macro & 42.2 & 50.9 & 52.4 & \best{59.4} & 64.3 & 63\% \\
\bottomrule
\end{tabular}
\end{table}

Per task (Table~\ref{tab:per-task}), the oracle-supervised critic closes 87\% of the
gap on position and 71\% on color attribution. Colors (22\%) and counting (41\%) lag behind. We hypothesize that color performance lags due to limited training data, and counting because the editor rarely succeeds (Figure~\ref{fig:editor-gap}).

\paragraph{A different-family critic.} Molmo reaches 52.4\% by reporting visual observations that a rule-based script converts into instructions. It makes more repairs than the baseline critic (301 vs.\ 226) but nearly three times the regressions (64), and paired against it the difference is not significant (201 vs.\ 167 images, $p=0.09$). It is the best zero-shot critic on counting (45.0\%), consistent with Molmo's pointing and counting training, and nearly matches the oracle-supervised Qwen critic on position (28.5\%). Trained on the same oracle labels with QLoRA, Molmo reaches 56.5\% with 345 repairs and 12 regressions (Table~\ref{tab:main}), 4.1 points above its zero-shot version with one fifth of the regressions. It recovers 65\% of the oracle's gain, against 78\% for the oracle-supervised Qwen critic, so oracle supervision improves critics from both model families.

\subsection{Generalization}
\label{sec:generalization}

\begin{table}[t]
\centering
\caption{Results on SD3.5-Large (downscaled to $512^2$) with the same VAREdit editor. The
oracle-supervised critic was trained only on SD1.5 images. $p$: exact McNemar test against the
original images. Rep./Reg.: repairs and regressions. Best non-oracle results in bold.}
\label{tab:sd35}
\small
\setlength{\tabcolsep}{3.5pt}
\begin{tabular}{lrrrrl}
\toprule
Instructions from & Macro & Micro & Rep. & Reg. & $p$ \\
\midrule
None (SD3.5-L) & 72.0 & 70.9 & -- & -- & -- \\
Baseline (Qwen) & 74.1 & 73.1 & 59 & \best{11} & $4{\times}10^{-9}$ \\
\textcolor{ours}{\textbf{Ours}} (oracle-sup.) & \best{76.0} & \best{75.1} & \best{109} & 16 & $3{\times}10^{-18}$ \\
\midrule
Oracle (reference) & 81.2 & 80.3 & 209 & 0 & $2{\times}10^{-63}$ \\
\bottomrule
\end{tabular}
\end{table}

\paragraph{Cross-generator transfer.} SD3.5-Large also exhibits an instruction gap: oracle instructions improve GenEval by 9.1 points (from 72.0\% to 81.2\%), repairing 209 of 644 failed images (Table~\ref{tab:sd35}). The oracle-supervised critic, trained only on SD1.5 images, transfers with a gain of 4.0
points, 109 repairs and 16 regressions (1.0\% of correct images). This is 44\% of the oracle's gain,
compared with 78\% on SD1.5. Position transfers best (62\% of the oracle's gain) and counting worst
(28\%; Appendix~\ref{app:checkpoints}). Paired against the baseline critic, the \ours\ repairs 71 images that the baseline critic does not,
against 26 in the other direction ($p=5.5\times10^{-6}$), and it closes 27\% of the instruction gap
on SD3.5 (1.9 of 7.1 points), compared with 63\% on SD1.5. Because the critic never saw an SD3.5 image during training or selection, this indicates that the error diagnosis learned from SD1.5 oracle labels transfers across generators, although by a smaller margin.
\paragraph{Unseen objects.} On the 1,428 official images whose objects never appear in the
critic's training prompts, the \ours\ closes 63\% of the instruction gap (59.1\% against 50.3\%
for the baseline critic and 64.4\% for the oracle; 173 vs.\ 53 paired repairs,
$p=5\times10^{-16}$), the same share as on all images; per-split results are in
Appendix~\ref{app:split}.

\paragraph{Human verification.} Because the critic is trained on and evaluated by the same detector, one annotator, blind to the method, evaluates randomly sampled repairs for prompt satisfaction and visual artifacts. On SD1.5, 72 of 100 repairs (50 oracle, 50 ours) are confirmed (95\% Wilson
interval 62--80\%); on SD3.5, 58 of 60 (89--99\%). A second SD1.5 sample of 40 repairs per method
confirms 80\% of the baseline critic's, 60\% of ours and 70\% of the oracle's ($p=0.09$, ours vs.\
baseline, Fisher's exact test); 12 of our 16 rejected repairs are color-attribution or position
prompts, where our critic gains most. 
Scaling repairs by human confirmation rates, minus regressions, yields 158 estimated net corrections for the baseline, 234 for ours, and 358 for the oracle. Our gain is 3.4 micro points versus GenEval's 9.0, though small samples limit confidence. Nearly all sampled SD3.5 repairs are confirmed.

\begin{table}[t]
\centering
\caption{Re-scoring GenEval images with T2I-CompBench's judges (SD1.5, VAREdit, 2,016 images;
scores $\times 100$). Color: GenEval's colors and color attribution; object: single and two objects
(BLIP-VQA); spatial: position; numer.: counting (UniDet). Mean: unweighted mean of the four columns.
Oracle instructions come from GenEval, not from these judges. Best results in bold.}
\label{tab:crossjudge}
\small
\setlength{\tabcolsep}{4pt}
\begin{tabular}{lrrrrr}
\toprule
Instructions & Color & Object & Spatial & Numer. & Mean \\
\midrule
None (original) & 50.75 & 68.63 & 1.46 & 68.95 & 47.45 \\
Baseline critic & 61.68 & 79.80 & 3.53 & 70.56 & 53.89 \\
\textcolor{ours}{\textbf{Ours}} & 75.81 & 85.52 & \best{10.36} & \best{72.98} & 61.17 \\
\midrule
Oracle (GenEval) & \best{77.07} & \best{86.58} & 10.32 & 72.18 & \best{61.54} \\
\bottomrule
\end{tabular}
\end{table}

\paragraph{A second editor.} With FLUX.1~Kontext [dev] and the same instruction files
(Table~\ref{tab:kontext}), the instruction gap is unchanged: oracle instructions reach 64.9\% and
the baseline critic 51.6\%, a gap of 13.3 points (13.4 with VAREdit). Our critic reaches 61.5\% and
closes 74\% of it (63\% with VAREdit); paired against the baseline critic, 294 images pass only under
ours and 65 only under the baseline ($p=7\times10^{-36}$). The editor gap is also nearly unchanged
(60.6\% of failures unrepaired with oracle instructions, against 61.3\%), so both gaps persist
across editors.

\begin{table}[t]
\centering
\caption{A second editor on SD1.5 (GenEval, same 2,212 images and instruction files). Rep./Reg.:
repairs and regressions against the original images.}
\label{tab:kontext}
\small
\setlength{\tabcolsep}{4pt}
\begin{tabular}{llrrrr}
\toprule
Editor & Instructions & Macro & Micro & Rep. & Reg. \\
\midrule
VAREdit & Baseline & 50.9 & 49.3 & 226 & 23 \\
VAREdit & Ours & 59.4 & 58.3 & 420 & 18 \\
VAREdit & Oracle & 64.3 & 63.3 & 512 & 0 \\
\midrule
FLUX.1 Kontext & Baseline & 51.6 & 49.9 & 231 & 15 \\
FLUX.1 Kontext & Ours & 61.5 & 60.3 & 450 & 5 \\
FLUX.1 Kontext & Oracle & 64.9 & 63.7 & 522 & 0 \\
\bottomrule
\end{tabular}
\end{table}

\paragraph{Independent judges.} GenEval's detector defines our labels, oracle, checkpoint selection
and main evaluation. We therefore re-score the same SD1.5 images with the judges of
T2I-CompBench~\citep{huang2023t2icompbench}, which play no part in training or selection: BLIP-VQA for
colors, color attribution and object presence, and UniDet for position and counting. Prompts are
rephrased into the judges' format (e.g., ``on the left of''); counting covers the 31 prompts whose
objects the judge knows, for 2,016 images in total. As scores are continuous, we compare images whose
score rises or falls with an exact sign test.

The ranking holds (Table~\ref{tab:crossjudge}): our critic raises the mean from 47.45 to 61.17,
against 53.89 for the baseline critic (527 vs.\ 257 images, $p=2.9\times10^{-22}$), and recovers 97\%
of the oracle's gain (78\% under GenEval). Against the original images, the baseline critic improves
382 images and worsens 101, ours 650 and 87, and the oracle 716 and 88 (all $p<10^{-38}$). The oracle's lead shrinks from 4.89 points to 0.37, which
suggests that part of it comes from fixes specific to GenEval's detector. Spatial scores are low for
every method because UniDet needs exact object names. On T2I-CompBench's own 900 prompts, our
critic also raises the judges' mean from 31.3 to 45.8, against 35.7 for the baseline critic
($p=3\times10^{-27}$, paired; Appendix~\ref{app:compbench}). These are new images from prompts outside GenEval's templates.

\section{Conclusion}

In this paper, we introduced LEDGE, which separates the two bottlenecks of text-to-image
self-correction by holding the images and the editor fixed and compiling oracle instructions from
the benchmark's requirements. Oracle instructions repair far more failures than the baseline critic,
yet 61\% of SD1.5 and 68\% of SD3.5 failures remain unrepaired under oracle instructions. The same oracle that exposes the instruction gap also helps close it. Using its labels as free supervision, we train a critic that improves exact diagnosis and recovers most of the instruction gap with fewer regressions. Trained only on SD1.5 images, the oracle-supervised critic also transfers to SD3.5-Large and to a second editor.

Given structured requirements and a verifier that identifies failures, LEDGE can evaluate different generators, critics, and editors. Correction methods
should report repairs and regressions next to an oracle reference: on SD1.5, a generic instruction
repairs more failures than the baseline critic, whose advantage comes from abstaining on correct
images. The two gaps also need different remedies: supervision compiled from the verifier narrows
the instruction gap, whereas the editor gap needs better editors.

Our labels, checkpoint selection and main evaluation share GenEval's detector. The ranking of
critics holds under T2I-CompBench's judges and on its free-form prompts, and the gain holds on the
65\% of images whose objects never appear in training, but the critic is trained with a single seed. Correction also adds an 8B critic
and an 8B editor, so we make no matched-compute claim against larger generators. Future work could train editors for counting and position, where the editor gap is largest, and distill verified corrections into the generator.

\section{LLM Usage Statement}
OpenAI's ChatGPT was used to assist with manuscript drafting, language refinement, grammar
checking, and presentation clarity. Anthropic's Claude was used to assist with writing experiment and analysis scripts (data preparation, evaluation, statistical tests, and figure generation). All experiments were carried out by the authors, who reviewed and verified all text, code, and results and remain responsible for scientific claims, experimental results, and conclusions.

{\small
\bibliographystyle{ieeenat_fullname}
\bibliography{references}
}

\clearpage
\setcounter{page}{1}
\maketitlesupplementary
\appendix

\section{Generator Baselines}
Table~\ref{tab:generators} reports official GenEval scores for every generator at native resolution
and for the $512^2$ SD3.5-Large copy used for correction.

\begin{table*}[t]
\centering
\caption{Official GenEval (553 prompts $\times$ 4 samples) for each generator at native resolution.}
\label{tab:generators}
\small
\begin{tabular}{lrrrrrrrr}
\toprule
Generator & Macro & Micro & Single & Two obj. & Count & Colors & Position & Color attr. \\
\midrule
SD1.5 & 42.23 & 40.14 & 95.94 & 34.34 & 37.19 & 73.67 & 4.50 & 7.75 \\
SDXL & 54.36 & 52.94 & 99.38 & 74.24 & 39.06 & 83.24 & 11.00 & 19.25 \\
FLUX.1-dev & 66.75 & 65.19 & 98.75 & 81.06 & 74.06 & 77.39 & 21.25 & 48.00 \\
SD3.5-Large & 70.99 & 69.85 & 99.06 & 89.65 & 70.00 & 83.24 & 26.00 & 58.00 \\
\midrule
SD3.5-Large, $512^2$ copy & 72.02 & 70.89 & 99.06 & 89.90 & 71.88 & 84.04 & 27.25 & 60.00 \\
\bottomrule
\end{tabular}
\end{table*}

\section{Macro and Micro Scores}
\label{app:macro}

Repairs and regressions convert directly to micro: for the oracle-supervised critic on SD1.5,
$888+420-18=1{,}290$ correct images, or $1{,}290/2{,}212=58.32\%$. Macro averages the per-task
changes. The net 402 images split as $+6, +140, +21, +39, +101, +95$ across single object, two
objects, counting, colors, position and color attribution (task sizes 320, 396, 320, 376, 400, 400),
changing those accuracies by $+1.88, +35.35, +6.56, +10.37, +25.25, +23.75$ points; their mean,
$+17.19$, gives $42.23+17.19=59.42\%$. Macro rises less than micro because most repairs fall in the
large tasks, whose images carry slightly less weight in the macro average.

\section{Instruction Templates}
\label{app:templates}

Every instruction ends with ``Preserve all unrelated objects, their attributes, and the background.''
Sentences are emitted in the order additions, removals, recolors, moves.

\begin{center}
\small
\begin{tabularx}{\linewidth}{lX}
\toprule
Edit & Template (example) \\
\midrule
add (none present) & Add one blue pizza in a clear, visible area of the scene. \\
add (some present) & Add two more handbags in a clear, visible area of the scene, leaving exactly four handbags. \\
add (relational) & Add one baseball glove to the right of the bear. \\
add (both missing) & Add one baseball glove and one bear, with the baseball glove to the right of the bear. \\
remove & Remove two extra handbags, leaving exactly four handbags. \\
recolor & Change the color of the pizza to blue. \\
move & Move the baseball glove to the right of the bear[, leaving a clear gap between them]. \\
\bottomrule
\end{tabularx}
\end{center}

\section{Structured Critique Schema}
\label{app:schema}

The structured critic receives the image, the rules below and the line ``Original prompt: $p$'', and returns
JSON.

\begin{figure*}[t]
\footnotesize
\begin{verbatim}
{"decision": "keep" | "edit", "edits": [ ... ]}   # keep <=> edits is empty
{"type": "add", "object", "color"|null, "count_to_add", "total",
 "relation"|null, "reference"|null, "reference_color"|null, "reference_missing"}
{"type": "remove", "object", "count_to_remove", "total"}
{"type": "recolor", "object", "color", "total"}
{"type": "move", "object", "relation", "reference", "gap"}
\end{verbatim}
\caption{Structured critique schema used by the structured zero-shot and oracle-supervised critics.}
\end{figure*}

Unparseable or invalid outputs become \textsc{review} (original image kept), the same policy as the baseline critic. On official SD1.5 images the oracle-supervised critic produced no invalid outputs.

\section{Training Data and Hyperparameters}
Table~\ref{tab:labels} breaks down the oracle labels on the pilot training images by task, and
Table~\ref{tab:hparams} lists the fine-tuning hyperparameters.

\begin{table}[t]
\centering
\caption{Oracle labels on pilot training images (SD1.5, corrected metadata).}
\label{tab:labels}
\small
\setlength{\tabcolsep}{4pt}
\begin{tabular}{lrrrr}
\toprule
Pilot tag & Images & Keep & Edit & Two edits \\
\midrule
Position & 448 & 43 & 405 & 0 \\
Counting & 376 & 37 & 339 & 175 \\
Color attribution & 322 & 28 & 294 & 103 \\
Color + position & 272 & 4 & 268 & 89 \\
Colors & 74 & 62 & 12 & 0 \\
Two objects & 38 & 17 & 21 & 0 \\
\midrule
Total & 1,530 & 191 & 1,339 & 367 \\
\bottomrule
\end{tabular}
\end{table}

\begin{table}[t]
\centering
\caption{Critic fine-tuning hyperparameters.}
\label{tab:hparams}
\small
\begin{tabularx}{\linewidth}{lX}
\toprule
Base model & Qwen3-VL-8B-Instruct (bf16), vision encoder frozen \\
LoRA & rank 16, $\alpha=32$, dropout 0.05; q, k, v, o, gate, up, down of the language model \\
Optimization & AdamW, lr $10^{-4}$, no weight decay, 5\% warmup then cosine, grad-norm clip 1.0 \\
Batch & 1 example, 8 accumulation steps \\
Data & 1,530 labeled images; correct images repeated $3\times$ (1,912 examples) \\
Schedule & 3 epochs, 717 optimizer steps; adapters saved every 50 steps \\
Selection & Best DEV exact match with $\le$10\% of correct DEV images edited $\rightarrow$ step 400 \\
Decoding & Greedy, up to 512 new tokens \\
\bottomrule
\end{tabularx}
\end{table}

\section{Checkpoint Selection and Per-Task Figures}
\label{app:checkpoints}

\begin{figure}[t]
\centering
\includegraphics[width=\linewidth]{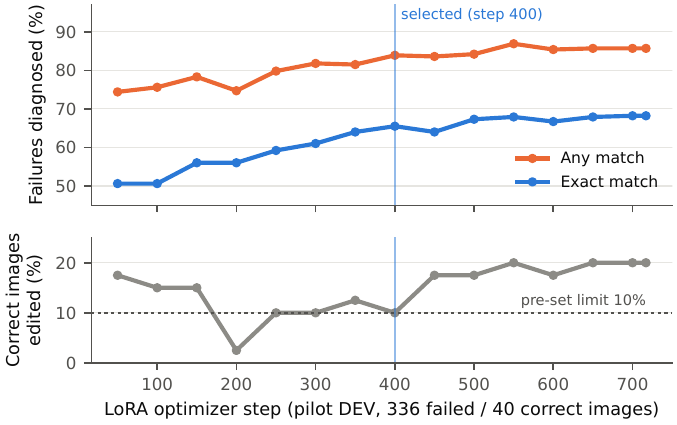}
\caption{Checkpoint selection on pilot DEV. Diagnosis improves steadily with training (top), but
after step 400 the oracle-supervised critic edits more correct images (bottom). The rule fixed in
advance selects step 400. With 40 correct DEV images, differences of one or two images are within
noise.}
\label{fig:checkpoints}
\end{figure}

\begin{figure}[t]
\centering
\includegraphics[width=\linewidth]{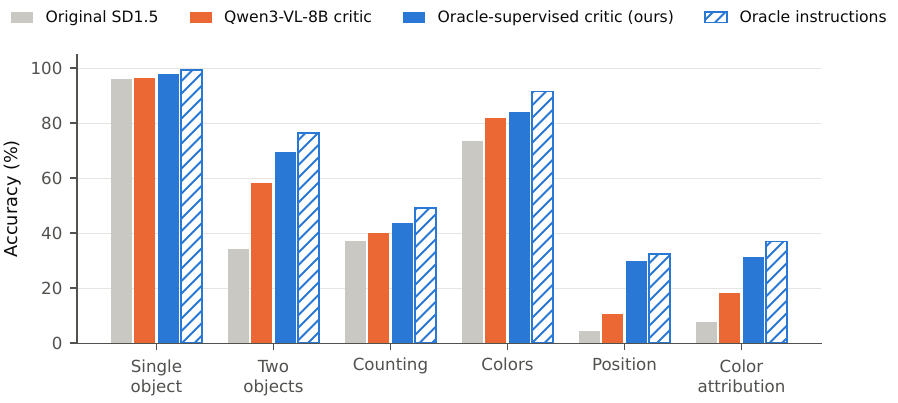}
\caption{Per-task GenEval accuracy on SD1.5 after editing, same editor (the data of
Table~\ref{tab:per-task}).}
\label{fig:per-task}
\end{figure}

\begin{figure}[t]
\centering
\includegraphics[width=\linewidth]{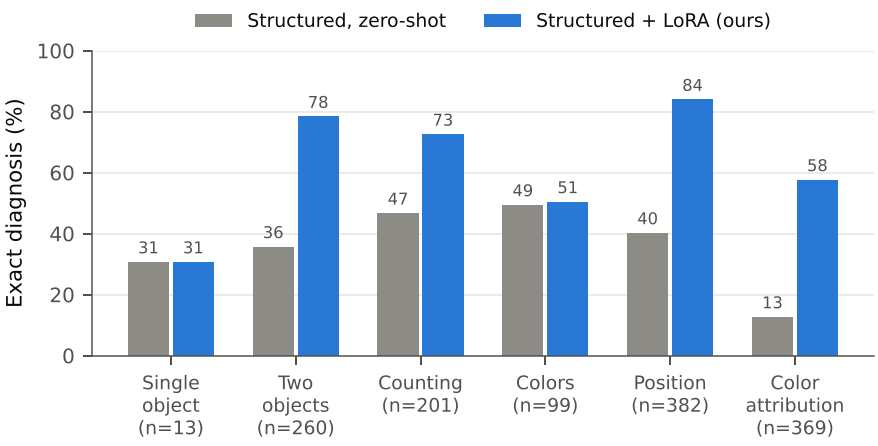}
\caption{Exact diagnosis by task on official SD1.5 failures. Training most improves color
attribution, position and two-object diagnosis. Colors improves little; it is the task least
represented in training (37 of 765 pilot prompts).}
\label{fig:diagnosis}
\end{figure}

\begin{figure}[t]
\centering
\includegraphics[width=\linewidth]{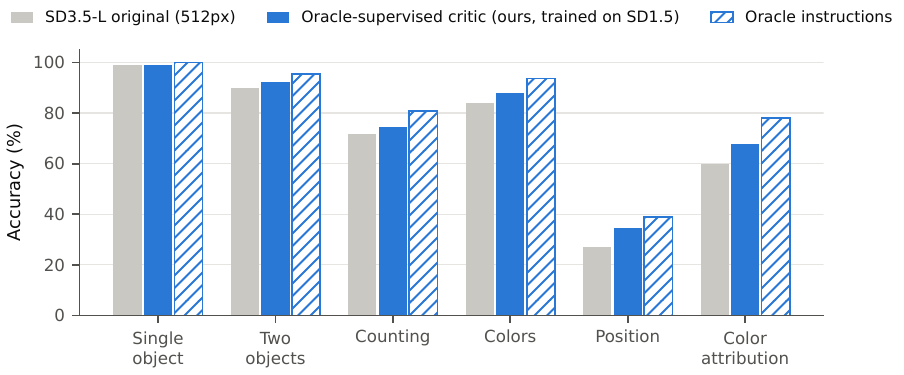}
\caption{Per-task accuracy on SD3.5-Large at 512px. The oracle-supervised critic transfers from
SD1.5 without retraining, recovering 44\% of the oracle's gain; position transfers best (62\%).}
\label{fig:sd35}
\end{figure}

Figure~\ref{fig:checkpoints} shows the trade-off that motivates the pre-declared rule. Exact
diagnosis on DEV rises from 50.6\% at step 50 to 68.2\% at step 717, but the share of correct images
edited rises to 20\% after step 400, since 87.5\% of the training labels are edits. The rule selects
step 400 (65.5\% exact, 4 of 40 correct images edited). Figures~\ref{fig:per-task},
\ref{fig:diagnosis} and~\ref{fig:sd35} plot the per-task results discussed in Section~\ref{sec:critic}
and Section~\ref{sec:generalization}.

\section{T2I-CompBench Prompts}
\label{app:compbench}

We also generate two SD1.5 images for each of the 900 validation prompts of T2I-CompBench's color,
spatial and numeracy categories and correct them with the same critics and editor (VAREdit). These
free-form prompts are written in a different style from GenEval's templates, and no oracle label or
GenEval verdict is involved: the critics are applied as trained, and the images are scored only by
T2I-CompBench's judges (Table~\ref{tab:compbench}).

\begin{table}[h]
\centering
\caption{T2I-CompBench's own prompts (SD1.5, VAREdit, 1,800 images; scores $\times100$). Against
the original images, the baseline critic improves 325 images and worsens 142 ($p=1.4\times10^{-17}$,
exact sign test) and ours 636 and 159 ($p=2.7\times10^{-68}$); paired, ours improves 577 images
over the baseline critic and worsens 266 ($p=3.4\times10^{-27}$). Best results in bold.}
\label{tab:compbench}
\small
\setlength{\tabcolsep}{4pt}
\begin{tabular}{lrrrr}
\toprule
Instructions & Color & Numer. & Spatial & Mean \\
\midrule
None (original) & 36.87 & 45.50 & 11.43 & 31.27 \\
Baseline critic & 48.88 & 45.89 & 12.27 & 35.68 \\
\textcolor{ours}{\textbf{Ours}} & \best{66.38} & \best{52.73} & \best{18.40} & \best{45.83} \\
\bottomrule
\end{tabular}
\end{table}

\section{Resampling Instead of Editing}
\label{app:resample}

We select 150 GenEval prompts at random, stratified by task, and keep all their samples (600 SD1.5
images, 374 failures). Each image is regenerated once with SD1.5 and a new seed and scored with
GenEval. A \emph{resample} variant replaces an image by its regeneration wherever a method would edit
it: every failure for the oracle-flagged variant, and every \textsc{edit} decision for the critics.
Edited results are the VAREdit outputs of the main experiments on the same images
(Table~\ref{tab:resample}). Regeneration alone, even when the oracle selects which images to
regenerate, stays below editing with the oracle-supervised critic, except on counting.

\begin{table*}[t]
\centering
\caption{Resampling vs.\ editing on a 600-image SD1.5 sample (374 failures). Rep./Reg.: repairs and
regressions against the original images. Right: share of failures repaired per task (number of
failures in parentheses).}
\label{tab:resample}
\small
\setlength{\tabcolsep}{3.5pt}
\begin{tabular}{lrrrrrrrrrr}
\toprule
Method & Macro & Micro & Rep. & Reg. & Single (6) & Two obj. (75) & Count. (58) & Colors (30) & Pos. (106) & C.\,attr. (99) \\
\midrule
None (original) & 39.7 & 37.7 & -- & -- & -- & -- & -- & -- & -- & -- \\
Oracle-flagged resample & 49.7 & 47.7 & 60 & 0 & 66.7 & 26.7 & \best{25.9} & 40.0 & 2.8 & 6.1 \\
Baseline critic, resample & 43.6 & 41.5 & 33 & 10 & 0.0 & 13.3 & 22.4 & 20.0 & 0.9 & 3.0 \\
Ours, resample & 45.8 & 43.8 & 43 & 6 & 33.3 & 22.7 & 22.4 & 13.3 & 2.8 & 4.0 \\
\midrule
Baseline critic, edit & 49.0 & 47.5 & 67 & 8 & 33.3 & 42.7 & 12.1 & 40.0 & 5.7 & 8.1 \\
Ours, edit & 57.6 & 56.7 & 118 & 4 & 50.0 & 57.3 & 10.3 & 40.0 & 32.1 & \best{20.2} \\
Oracle, edit & \best{61.8} & \best{60.7} & \best{138} & 0 & 66.7 & \best{61.3} & 22.4 & \best{63.3} & \best{34.9} & 19.2 \\
\bottomrule
\end{tabular}
\end{table*}

\section{Editor Failures}
\label{app:editor-failures}

For every failure that receives an oracle edit, we parse GenEval's failure reasons before and after
editing with the oracle's own parser and compare them per object. A requirement is \emph{not
executed} if the same reason remains (the object is still missing, extra, miscolored or misplaced)
and \emph{mis-executed} if the object now fails differently (e.g., added on the wrong side or in the
wrong color). An \emph{unintended change} is a new failure on an object that satisfied the prompt
before editing. GenEval sees only the prompt's objects, counts, colors and relations, so other
changes to the scene are not counted.

\begin{table}[h]
\centering
\caption{Unrepaired images under oracle instructions (VAREdit), by failure category. Categories
overlap; percentages are of unrepaired images.}
\label{tab:editor-failures}
\small
\begin{tabular}{lrr}
\toprule
Category & SD1.5 & SD3.5 \\
\midrule
Unrepaired / failures & 812 / 1,324 & 435 / 644 \\
Failed insertion & 56.3\% & 29.0\% \\
Incorrect attribute modification & 17.5\% & 14.7\% \\
Failed move & 18.0\% & 48.7\% \\
Unintended change & 9.5\% & 7.6\% \\
\bottomrule
\end{tabular}
\end{table}

\section{Reproducibility}

\begin{sloppypar}
All scripts are in the project repository: oracle compilation (\texttt{oracle\_instructions.py}),
critic schema and rendering (\texttt{critic\_schema.py}), critic training and inference
(\texttt{train\_critic.py}, \texttt{critic\_infer.py}), diagnosis scoring (\texttt{score\_critic.py}),
paired comparison (\texttt{compare\_runs.py}), pilot metadata correction and re-scoring
(\texttt{fix\_pilot\_metadata.py}, \texttt{rescore\_tree.py}), resolution control
(\texttt{downscale\_run.py}), splits (\texttt{split\_by\_objects.py}, \texttt{subset\_by\_cases.py}),
the VAREdit runner (\texttt{benchmark.py}), the second-editor runner (\texttt{edit\_diffusers.py}),
the image controls (\texttt{image\_controls.py}, the resample
baseline (\texttt{small\_sample.py}, \texttt{run\_resample.sh}), the editor-failure analysis
(\texttt{editor\_failures.py}) and the T2I-CompBench re-scoring
(\texttt{export\_geneval\_for\_compbench.py}, \texttt{compare\_compbench.py}).
Generator runs record per-image SHA-256 checksums; the SD1.5 run manifest hash is
\texttt{2ebbcbd7\ldots}. Model revisions are pinned in \texttt{model\_lock.json}; the Qwen3-VL commit
used for training is recorded in the adapter configuration.
\end{sloppypar}
\section{Results by Object Split}
\label{app:split}
Images are split by whether their prompt's objects appear among the 20 classes of the training
prompts: none (unseen, 1,428 images), some (mixed, 464) or all (seen, 320).

\begin{table}[h]
\centering
\caption{GenEval macro accuracy by object split (SD1.5, VAREdit). Base.: baseline critic; Molmo:
oracle-supervised Molmo (QLoRA); closed: share of the baseline-to-oracle gap closed by the
oracle-supervised critic.}
\label{tab:split}
\small
\setlength{\tabcolsep}{3.5pt}
\begin{tabular}{lrrrrrr}
\toprule
Split & Orig. & Base. & Molmo & Ours & Oracle & Closed \\
\midrule
Unseen (1,428) & 41.7 & 50.3 & 55.3 & 59.1 & 64.4 & 63\% \\
Mixed (464) & 15.5 & 30.2 & 40.4 & 43.2 & 47.8 & 74\% \\
Seen (320) & 42.1 & 49.8 & 55.2 & 60.6 & 65.8 & 67\% \\
\midrule
All (2,212) & 42.2 & 50.9 & 56.5 & 59.4 & 64.3 & 63\% \\
\bottomrule
\end{tabular}
\end{table}

\end{document}